\documentclass[10pt,journal]{IEEEtran}
\usepackage{amssymb}
\usepackage{multirow,setspace,verbatim,amsfonts,graphicx,amsmath,amsthm,amsbsy,amssymb,epsfig,url}
\usepackage{amsthm}
\usepackage{gensymb}
\usepackage{booktabs}
\usepackage{cite}
\usepackage{makecell}
\usepackage{epstopdf}
\usepackage{algorithm}
\usepackage[noend]{algpseudocode}
\usepackage{amsfonts}
\usepackage{amsmath,bm}
\usepackage{amssymb}
\usepackage{amsthm}
\usepackage{tikz,graphicx}
\usepackage{mathrsfs} 
\usepackage[algo2e]{algorithm2e} 

\usepackage{array,multirow}
\usepackage{color}
\usepackage{caption}
\usepackage{subcaption}

\newcommand{\Bb}{{\mathbf B}}
\newcommand{\Cb}{{\mathbf{C}}}

\newcommand{\Gb}{{\mathbf G}}

\newcommand{\Ib}{{\mathbf I}}

\newcommand{\Pb}{{\mathbf P}}

\newcommand{\Ub}{{\mathbf U}}
\newcommand{\Vb}{{\mathbf V}}

\newcommand{\Yb}{{\mathbf Y}}
\newcommand{\Zb}{{\mathbf Z}}

\newcommand{\fb}{{\mathbf f}}
\newcommand{\gb}{{\mathbf g}}
\newcommand{\hb}{{\mathbf h}}

\newcommand{\kb}{{\mathbf k}}

\newcommand{\rb}{{\mathbf r}}

\newcommand{\ub}{{\mathbf u}}
\newcommand{\vb}{{\mathbf v}}

\newcommand{\xb}{{\mathbf x}}
\newcommand{\yb}{{\mathbf y}}
\newcommand{\zb}{{\mathbf z}}

\newcommand{\Phib}{{\boldsymbol {\Phi}}}
\newcommand{\Psib}{{\boldsymbol {\Psi}}}

\newcommand{\Rd}{{\mathbb R}}
\newcommand{\Cd}{{\mathbb C}}

\newcommand{\phib}{{\boldsymbol{\phi}}}
\newcommand{\psib}{{\boldsymbol{\psi}}}

\newcommand{\Hbc}{{\boldsymbol{\mathcal H}}}

\newcommand{\Kc}{{{\mathcal K}}}
\newcommand{\Fc}{{{\mathcal F}}}
\newcommand{\Pc}{{{\mathcal P}}}

\newcommand{\rank}{\textsc{rank}}
\newcommand{\hank}{\mathbb{H}}

\newcommand{\NN}{\mathbb{N}}
\newcommand{\RR}{\mathbb{R}}

\newcommand{\bk}{\mathbf{k}}

\newcommand{\beq}{\begin{equation}}
\newcommand{\eeq}{\end{equation}}
\newcommand{\beqa}{\begin{eqnarray}}
\newcommand{\eeqa}{\end{eqnarray}}

\begin{document}

\title{ $k$-Space Deep Learning for Parallel MRI:  \\ Application to 
Time-Resolved MR Angiography   %DCE MRI %Dynamic Contrast Enhanced MRI  %Dynamic Contrast Enhanced MRI Time-Resolved Angiography
}

\author{Eunju~Cha, Eung Yeop Kim,~%\IEEEmembership{Student Member,~IEEE,}
%        Junhong~Min,~%\IEEEmembership{Member,~IEEE,}
        and~Jong~Chul~Ye$^{*}$,~\IEEEmembership{Senior Member,~IEEE}% <-this % stops a space
\thanks{EJC and JCY are with the Department of Bio and Brain Engineering, Korea Advanced Institute of Science and Technology (KAIST), 
		Daejeon 34141, Republic of Korea (e-mail: \{eunju.cha,jong.ye\}@kaist.ac.kr). 
		EYK is with Department of Radiology, Gil Medical Center, Gachon University College of Medicine, Incheon, South Korea.
		This work is supported by National Research Foundation of Korea, Grant number NRF2016R1A2B3008104.}% <-this % stops a space
}		

\maketitle

\begin{abstract}

%\noindent\textbf{Purpose:} 
 Time-resolved angiography with interleaved stochastic trajectories (TWIST) has been widely used 
 for dynamic contrast enhanced MRI (DCE-MRI). To achieve highly accelerated acquisitions, 
 TWIST combines the periphery of the $k$-space data from several adjacent frames   to reconstruct one temporal frame.
 However, this view-sharing scheme limits the true temporal resolution of TWIST. 
Moreover, the $k$-space sampling patterns have been  specially designed  for a specific
generalized autocalibrating partial parallel acquisition (GRAPPA) factor so that it is not possible to reduce the number of view-sharing %later in order to reconstruct images with a better temporal resolution. 
once the  $k$-data is acquired.
To address these issues, this paper proposes a novel $k$-space deep learning approach for parallel MRI. 
In particular, %inspired by the recent mathematical discovery that links Hankel matrix decomposition to deep learning, 
we have designed our neural network so that accurate $k$-space interpolations are performed simultaneously for multiple coils by exploiting the redundancies along the coils and images.
%Thanks to the excellent generalization capability, 
Reconstruction results using in vivo TWIST data set confirm that
the proposed method can immediately generate high-quality reconstruction results with  various choices of view-sharing, allowing us to exploit the trade-off between spatial and temporal resolution
%
%Reconstruction results using in vivo TWIST data set confirm the excellent  generalization capability of the proposed $k$-space
%deep learning 
in time-resolved MR angiography. % method.
%that  the proposed $k$-space deep learning significantly improves temporal resolution while maintaining the similar spatial resolution to GRAPPA. 
\end{abstract}


\section{Introduction}

DCE-MRI is one of the important imaging protocols for clinical applications. In  DCE-MRI, a series of MR images are acquired after the injection of the contrast agent to patients. DCE-MRI provides information on the physiological characteristics of tissues, such as status of blood vessels, so it is useful for stroke or cancer imaging \cite{turnbull2009dynamic,yankeelov2009dynamic}.    
In particular, the time-resolved angiography with interleaved stochastic trajectories (TWIST) \cite{laub2006syngo} is widely used due to its superior temporal and spatial resolution. 

In TWIST, the center and the periphery of $k$-space data are acquired at different rates. Specifically, the low frequency region is completely sampled to retrain the overall image contrast, but the high frequency region is randomly sub-sampled at each time frame. Then, the high frequency regions of the $k$-space from multiple temporal frames are  combined to obtain uniformly sub-sampled $k$-space data so that the missing data can be interpolated using GRAPPA \cite{griswold2002generalized}. Thanks to the aggressive under-sampling of the periphery of $k$-space data, TWIST offers a significant improvement in both temporal and spatial resolution. It is known that TWIST imaging can follow the perfusion dynamics  more closely, which is useful for time-resolved MR angiography (tMRA)\cite{nael2009time}.

\begin{figure*}[!hbt] 	
\center{ 
\includegraphics[width=16cm]{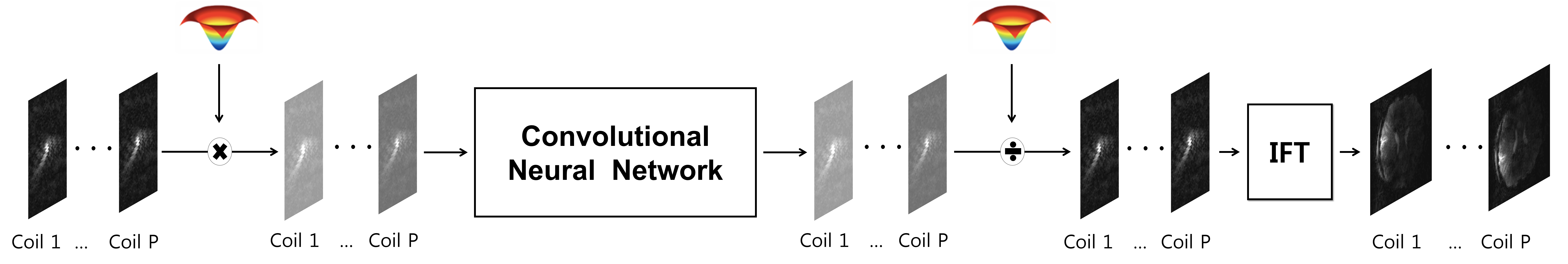}
}
\caption{The architecture of $k$-space deep learning for parallel MRI. Here, IFT stands for inverse Fourier transform.}
\label{fig:scheme}
\end{figure*}

However, the temporal resolution of TWIST is not a true one due to the extensive view-sharing from several adjacent frames, so the quantitative study of perfusion dynamics using TWIST alone is not usually recommended.
Another important disadvantage of TWIST is that the sampling pattern is designed for specific GRAPPA acceleration factor, so it is not possible to 
change the number of view sharing once the acquisition is done.

Despite the needs for new image reconstruction algorithms,
 there are several technical difficulties in developing methods to address these problems.
Since the $k$-space samples from the reduced view-sharing are
 a subset of uniformly subsampled $k$-space data, the sampling pattern is not incoherent, so the existing compressed sensing (CS) approaches\cite{jung2009k,lustig2007sparse} have difficulties in removing aliasing artifacts.
In our previous works \cite{cha2017true},   we therefore proposed to improve temporal resolution of  TWIST  via $k$-space interpolation 
using ALOHA \cite{7547372, lee2016acceleration, lee2016reference} that can synergistically combine parallel MRI (pMRI) and CS-MRI by exploiting the sparsity and the inter-coil redundancy in a unified matrix completion framework. However, the computational cost for TWIST reconstruction using ALOHA was very expensive due to the multiple matrix factorization to allow 4-dimensional TWIST imaging. Moreover, spatial resolution losses are often observed when the number of view sharing is not sufficient. Therefore, a new approach is required to overcome this limitation.

Recently, deep learning approaches have been extensively employed for computer vision applications thanks to the availability of  the massive datasets and high performance  graphical processing units (GPUs) \cite{krizhevsky2012imagenet, he2016deep}.
In MR literature,  the works in \cite{wang2016accelerating,hammernik2018learning,kwon2017parallel}  were among 
the first that applied deep learning approaches to CS MRI.  
%In particular,
%Deep network architecture using unfolded iterative compressed sensing (CS) algorithm was also proposed 
%\cite{hammernik2018learning}. 
 These works were followed by novel
extension using deep residual learning \cite{lee2018deep}, domain adaptation \cite{han2017deep}, data consistency layers \cite{schlemper2018deep}, etc.
All these pioneering works have consistently demonstrated superior reconstruction performances over the compressed sensing approaches \cite{lustig2007sparse,jung2009k,lingala2011accelerated,jin2016general,lee2016acceleration}
at significantly lower run-time computational complexity.

Therefore, the purpose of this research is to develop a deep learning approach that
can  improve the temporal resolution of TWIST imaging by reducing the number of view-sharing. Moreover,
we aim at developing algorithm that can reconstruct images at various choices of  view sharing  to see the trade-off between the spatial and temporal resolution.

However, the application of deep learning for TWIST requires overcoming two major technical hurdles.
First, to be backward compatible with TWIST imaging as well as to allow reconstruction with various number of view-sharing,  
the deep network is required to learn  $k$-space interpolation kernels.
% by exploiting the multi-coil redundancies and spatial domain redundancy. 
However,  most of the popular deep learning MR reconstruction algorithms
are either in the form of  image domain post-processing \cite{kwon2017parallel,lee2018deep,han2017deep}, or iterative updates between the $k$-space and the image domain using a cascaded network \cite{hammernik2018learning,wang2016accelerating,schlemper2018deep},
which are  different from GRAPPA-based TWIST protocol.
%An extreme form of the neural network called AUtomated TransfOrm by Manifold APproximation (AUTOMAP) \cite{zhu2018image}   as shown in Fig.~\ref{fig:deepMR}(a) even attempts to estimate the Fourier transform itself using fully connected layers.
Second, with reduced view sharing, standard GRAPPA fails to provide reasonable reconstruction quality,
so there are no ground-truth data that can be used as labels.

One of the main contributions of this paper is therefore to show that the recent $k$-space deep learning approach \cite{han2018k} is very versatile in meeting these technical challenges. More specifically,  the $k$-space deep learning \cite{han2018k} was inspired by the recent mathematical discovery
that  a deep convolutional neural network can be derived as a data-driven decomposition of  Hankel matrix so that a neural network can be effectively designed
 in the domain where Hankel matrix can have a  low-rank structure  \cite{ye2018deep}.
Recall that  the basic idea of ALOHA for parallel MRI \cite{jin2016general}
is based on the observation that the extended Hankel matrix in the $k$-space domain constructed by stacking Hankel matrices from each coil side-by-side has a low-rank structure. 
This implies that, in contrast to  the common practice that the neural networks are implemented in the image domain,
a better neural network should be constructed in the $k$-space domain by stacking multi-coil $k$-space data along the channel direction of the neural network. Then,  
 our deep neural network  is trained to learn the relationship between the multi-coil $k$-space channel data and the channel-by-channel reconstructed coil images as shown in Fig.~\ref{fig:scheme}.
%The link between ALOHA and deep learning also leads us how

To overcome the lack of ground-truth data for different temporal resolutions and allow flexible reconstruction for different numbers of view-sharing,   our neural network is designed to learn the $k$-space interpolation relationship between the minimum number of $k$-space samples and the fully sampled $k$-space data from the GRAPPA reconstruction.
%Another important discover is to show that 
Interestingly, the thus-trained neural network is generalized very well for all temporal resolution, since
 our neural network is to learn the Fourier domain structure rather than the image content.
%the thus-trained neural network can immediately generate reconstruction results with  different  spatial and temporal
%resolution by simply changing the number of 
%view sharing.

As a byproduct, our theory and numerical verification can address some of the fundamental issues in designing a deep neural network for MR reconstruction.
In particular, the current practice of splitting real and imaginary channels as well as multi-coil data is valid for $k$-space interpolation using neural network,
 because it preserves the low-rank nature of  Hankel-structured matrix.
In addition, our theoretical analysis confirm that we do not need to reinvent any new non-linearity for complex-valued MR image reconstruction problems because the main role of ReLU's positivity constraint is to allow a conic decomposition of the Hankel matrix.
The implication of conic decomposition will be detailed later.

%The structure of this paper is as follows.  In Theory, we first present image reconstruction problems for TWIST imaging with reduced view sharing.
%Then,  we review the theory of ALOHA for parallel imaging, which is followed by its extension to deep neural nework.
%The data preparation, neural network structure and training is explained in Methods, which is followed by Experimental Results, Discussions and Conclusions.

\section{Theory}
\subsection{Problem Formulation }

\begin{figure}[!b] 	
\center{ 
\includegraphics[width=7cm]{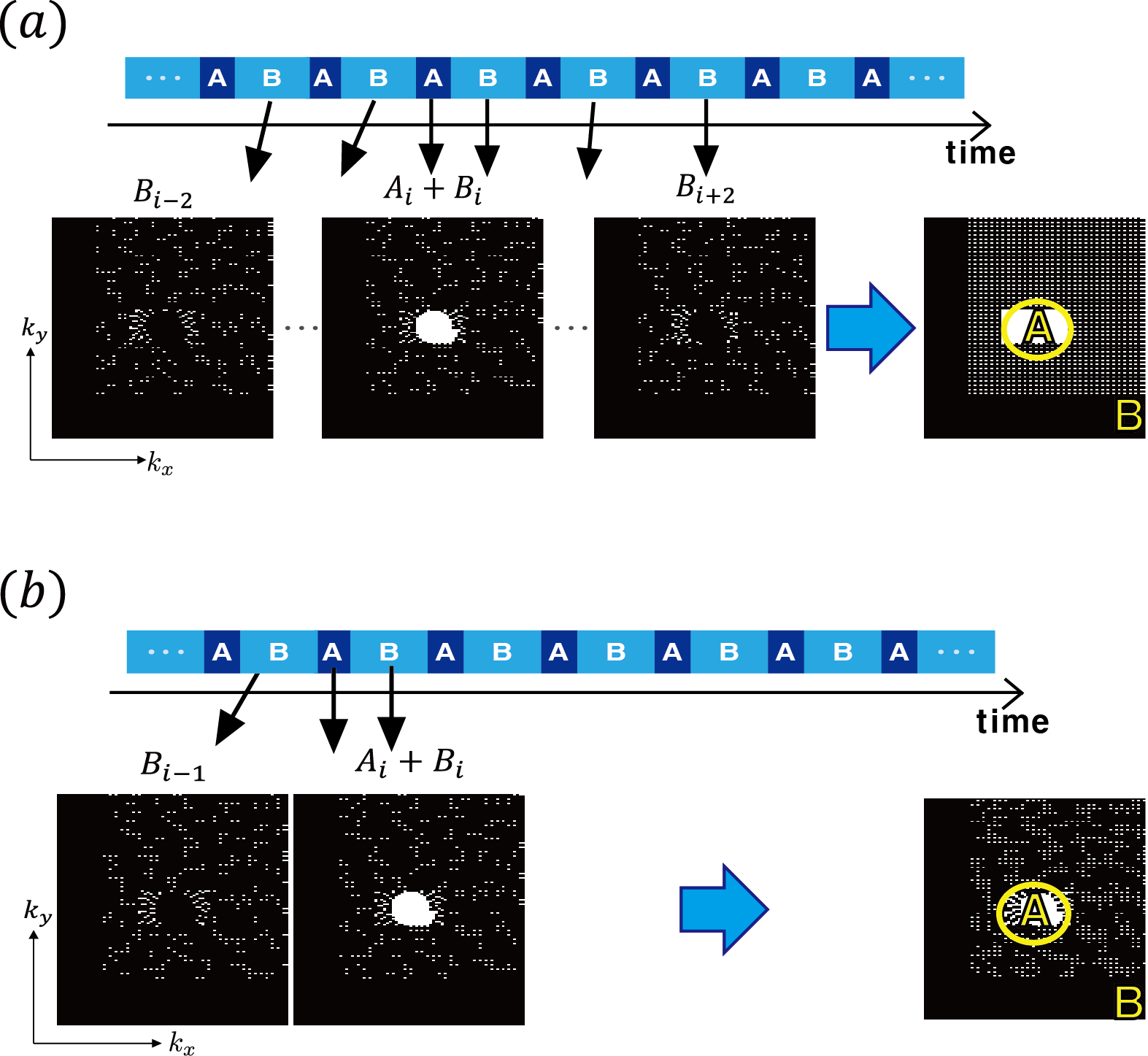}
}
\caption{View-sharing scheme for our carotid vessel data. The center and periphery of $k$-space are designated  A and B, respectively. (a) Conventional scheme for 2D GRAPPA reconstruction, and  (b) an example of  reduced view sharing.}
\label{fig:TWIST_concept}
\end{figure}

 In TWIST, the center and periphery of $k$-space data are sampled at different rates. More specifically, at each time frame, the center of $k$-space data (A region in Fig. \ref{fig:TWIST_concept}) is fully sampled while the periphery of $k$-space data (B region in Fig. \ref{fig:TWIST_concept}) is partially sampled. Thus, the acquired
 $k$-space samples  can be reduced for each frame, resulting in a reduced acquisition time. However, due to the strongly subsampled high frequency $k$-space data, individual reconstruction of each frame provides degraded images. Therefore, high frequency $k$-space data should be combined from adjacent frames to create a time frame. The predominant justification for this sliding window approach is that the image contrast is assumed to be dependent on the low-frequency  $k$-space center, so that the dynamics of the contrast agent are not altered by such view sharing.
 However, as shown in our experimental section, this common belief is not always true because the temporal dynamics of the contrast agents are degraded by sharing views. Therefore, the actual temporal resolution of TWIST imaging is determined by the number of view-sharing.

There are different types of TWIST sampling patterns. For example,  the one in Fig. \ref{fig:TWIST_concept}(a) was developed specifically for 2-D GRAPPA reconstruction,
where  high frequency regions of five time frames ($B_{i-2}, \cdots, B_{i+2}$) are integrated to generate a 2-D uniform sub-sampled $k$-space data  with the downsampling factor of three and two along $k_x$ and $k_y$ directions, respectively. Then, 2-D GRAPPA \cite{griswold2002generalized} is used to interpolate the missing elements in that $k$-space. Since the net sliding window corresponds to 9 frames (five B regions and four A
 frames), the resulting temporal resolution is severely impaired. 

Unlike the existing TWIST, which utilizes the five adjacent frames for the 2-D GRAPPA reconstruction, we are interested in using various number of
view-sharing.
For example, Fig. \ref{fig:TWIST_concept}(b) shows the number of view-sharing is reduced to two frames, which is considered the minimum number of view-sharing in this paper.
The reduced view-sharing results in highly under-sampled and  irregular sampling pattern that is difficult to apply the existing GRAPPA algorithm. 
Therefore, our goal is to develop a multi-coil deep learning approach to interpolate the missing $k$-space data.

To provide a mathematical formulation of our imaging problem,
the spatial Fourier transform of a function $x:\RR^2\to\RR$ is first defined by 
\begin{align*}
\hat{x}(\bk)=\mathcal{F}[x](\bk):=\int_{\RR^d} e^{-\iota\bk\cdot \rb}x(\rb)d\rb,
\end{align*}
with spatial frequency $\bk\in\RR^2$ and $\iota=\sqrt{-1}$.
Let  $\{\bk_n\}_{n=1}^N$, for some integer $N\in\NN$, be a collection of finite number of sampling points of the $k$-space
 confirming to the Nyquist sampling rate. 
 Accordingly, the discretized $k$-space data  $\widehat\xb\in \Cd^N$  is introduced by
\begin{equation}%\label{eq:coil}
%\widehat \Xb := \begin{bmatrix} \widehat \xb_1 & \cdots & \widehat \xb_C \end{bmatrix} \in \Cd^{N\times C} \quad\text{with}\quad
\widehat \xb = \begin{bmatrix} \widehat x(\kb_1) & \cdots  & \widehat x(\kb_N) \end{bmatrix}^T   \  .
\end{equation}
In parallel MRI, the unknown 2-D image $g_i(\rb), \rb\in \RR^2$ from the $i$-th coil  can be represented as
\beq\label{eq:gi}
g_i(\rb) = s_i(\rb) x(\rb),\quad i=1,\cdots, P, 
\eeq
where $s_i(\rb)$ denotes the $i$-th coil sensitivity map, $x(\rb)$ is an unknown image, and $P$ denotes the number of coils.
Then, the $k$-space data at the $i$-th coil 
is defined by 
\begin{align*}
\widehat g_i(\bk)= \mathcal{F}[g_i](\bk) %:= \int_{\RR^d} e^{-\iota\bk\cdot \rb}s_i(\rb)x(\rb)d\rb,
\end{align*}
whose  discretized $k$-space data  $\widehat\gb_i\in \Cd^N$  is denoted by
\begin{equation}\label{eq:coil}
%\widehat \Xb := \begin{bmatrix} \widehat \xb_1 & \cdots & \widehat \xb_C \end{bmatrix} \in \Cd^{N\times C} \quad\text{with}\quad
\widehat \gb_i = \begin{bmatrix} \widehat g_i(\kb_1) & \cdots  & \widehat g_i(\kb_N) \end{bmatrix}^T   \  .
\end{equation}
  For a given under-sampling pattern $\Lambda$ from the reduced view sharing, let 
  the downsampling operator $\Pc_\Lambda: \Cd^{N} \to \Cd^{N}$ 
  be defined as
  \begin{eqnarray}
  \left[\Pc_\Lambda[\hat \xb] \right]_j= \begin{cases} \left[\widehat\xb\right]_j &j \in \Lambda \\
0, &  \mbox{otherwise} \end{cases}   \   .
  \end{eqnarray}
The main goal of parallel imaging is then to exploit the common signal $x(\rb)$ that is measured through multiple channels.
Specifically, our image reconstruction problem   is given by
\begin{eqnarray}\label{eq:fwd}
\min_{x,\{g_i\}_{i=1}^{P}} & &R\left(x,\{g_i\}_{i=1}^{P}\right) \\
\mbox{subject to} &&\hat \yb_i  :=\Pc_\Lambda[\hat \gb_i] ,\quad i=1,\cdots, P
\end{eqnarray}
where $R\left(x,\{g_i\}_{i=1}^{P}\right)$ denotes some regularization term that depends on the unknown signal $x(\rb)$
and coil images $g_i(\rb),i=1,\cdots, P$. In CS-MRI  \cite{lustig2007sparse,jung2009k,lingala2011accelerated},
the regularization term is usually chosen to have minimum non-zero support in some sparsifying
transform domain.

\subsection{Multi-coil $k$-Space Low-Rank Hankel Matrix}

Although ALOHA \cite{jin2016general,lee2016acceleration} also takes advantages of the image domain spasifying transform as in the conventional CS-MRI algorithms,
in contrast to the CS-MRI approaches,    ALOHA is  for direct
$k$-space interpolation.
Specifically, according to the theory of ALOHA \cite{ye2017compressive,jin2016general},
 if the underlying signal  in the image domain is sparse and  described as the signal with the finite rate of innovations (FRI) \cite{vetterli2002sampling},
  the associated Hankel matrix from its $k$-space data 
 is low-ranked.
 Therefore, if some of $k$-space data  are missing,
we can construct an appropriate weighted Hankel matrix with missing elements such that the missing elements are recovered 
using low rank Hankel matrix completion approaches \cite{candes2009exact,gross2011recovering}.
Beside  the low-rank property originating from  sparsity in the transform domain,  there exists an additional low-rank relationship that is unique in parallel MRI.
More specifically, Eq.~\eqref{eq:gi} leads to the following inter-coil relationship:
$$g_i(\rb)s_j(\rb) - g_j(\rb)s_i(\rb) = 0, \quad \forall i\neq j, $$
which is equivalent to the $k$-space inter-coil annihilating filter relationship \cite{jin2016general}:
\begin{eqnarray}\label{eq:intercoil}
\widehat g_i \circledast \widehat s_j - \widehat g_j\circledast \widehat s_i = 0,\quad \forall i\neq j . 
\end{eqnarray}
In \cite{jin2016general}, we formally showed that
the inter-coil annihilating filter relationship in \eqref{eq:intercoil} leads to the low-rank property of the
following extended Hankel structured matrix:
\begin{eqnarray}\label{eq:PHankel}
\hank_{d|P}(\widehat\Gb) = \begin{bmatrix} \hank_d(\widehat\gb_1) & \cdots & \hank_d(\widehat\gb_P) \end{bmatrix}
\end{eqnarray}
where $$\widehat\Gb=\begin{bmatrix} \widehat\gb_1 & \cdots & \widehat\gb_P \end{bmatrix} \in \Cd^{N\times P}$$
with the $k$-space measurement $\widehat \gb_i$ in \eqref{eq:coil}, and 
 $\hank_d(\widehat \gb_i)$ denote a 
 Hankel matrix constructed from $\widehat\gb_i$ with  $d$ denoting the
 matrix pencil size. For more details on the construction of Hankel matrices and their relation to the convolution, see Appendix A in the Supplementary Material.

 Therefore, if some of $k$-space data  are missing,
the missing elements are recovered 
using low rank Hankel matrix completion approaches \cite{candes2009exact,gross2011recovering}:
\begin{eqnarray}\label{eq:EMaC}
(MC)
 &\min\limits_{\widehat \Zb\in \Cd^{N \times P}} & \rank~ \hank_{d|P} (\widehat \Zb)  \\
&\mbox{subject to } & \Pc_\Lambda[\widehat\gb_i ] = \Pc_\Lambda[\widehat \zb_i]  ,\quad i=1,\cdots, P \nonumber  \  .
\end{eqnarray}
The low-rank Hankel matrix completion problem $(MC)$  can be solved in various ways, and ALOHA employs
the matrix factorization approaches  \cite{jin2016general,lee2016acceleration,lee2016reference}. 
%Previously, we demonstrated that the missing elements of $k$-space with reduced view-sharing can be  interpolated using ALOHA \cite{cha2017true}. 
However, the main technical issue is its relatively expensive computational cost for matrix factorization. In the following section, we show that a deep learning approach can address this problem by handling the matrix decomposition fully data-driven way.

\subsection{From ALOHA to Deep Neural Network} 

Recently, we showed that the Hankel matrix decomposition in ALOHA is closely related to deep neural network \cite{ye2018deep}.
To understand its link to a deep neural network, ALOHA optimization problem $(MC)$ is converted to the following regression problem:
\begin{eqnarray} 
\label{eq:image_regression}
(MC') \quad \quad\quad \min_{\widehat{\Zb} \in \Cd^{N\times P} }  && \sum_{i=1}^P\left\|g_i- \Fc^{-1}[\widehat{\zb}_i] \right\|^2 \\  
\mbox{subject to}&& \rank~ \hank_{d|P} (\widehat \Zb) = Q \label{eq:sol} \\
&&\Pc_\Lambda[\widehat\gb_i ] = \Pc_\Lambda[\widehat \zb_i]  ,\quad i=1,\cdots, P \nonumber  \  ,
\end{eqnarray}
where $Q$ is estimated rank of Hankel structured matrix and $\Fc$ denotes Fourier transform. Note that
 the low rankness is enforced in the $k$-space, whereas the cost function is defined as
 the image reconstruction error for each coil. %constraint are respectively defined are different.

Now, for any feasible solution $\widehat\Zb$ for \eqref{eq:sol}, suppose that the singular value decomposition of the associated Hankel structured matrix is given by
$\hank_{d|P}(\widehat\Zb)=\Ub \bold{\Sigma} \Vb^\top$, where $\Ub=[\ub_1~\cdots~\ub_Q] \in \Rd^{N\times Q}$ and $\Vb=[\vb_1~\cdots~\vb_Q] \in \Rd^{d\times Q}$ are the left and right singular vector bases matrices, respectively; $\bold{\Sigma} = (\sigma_{ij}) \in \Rd^{Q \times Q}$ is the diagonal matrix with singular values. Let $\Psib=[\psib_1,\cdots,\psib_Q]$ and $\widetilde{\Psib}=[\widetilde\psib_1,\cdots,\widetilde\psib_Q]$ $\in \Rd^{d \times Q}$ are a pair of matrix satisfying the low-dimensional subspace constraint:
\begin{eqnarray}\label{eq:projection}
\Psib \widetilde \Psib^{\top} = \Pb_{R(\Vb)} ,
\end{eqnarray}
where  $\Pb_{R(\Vb)}$ denotes the projection matrix to the range space of $\Vb$. Similarly, another pair of matrices $\Phib=[\phib_1,\cdots,\phib_M]$ and $\widetilde{\Phib}=[\widetilde\phib_1,\cdots,\widetilde\phib_M] \in \Rd^{N \times M}$
satisfy the so-called the frame  condition \cite{ye2018deep}:
\begin{eqnarray}\label{eq:projectionU}
\Phib \widetilde \Phib^{\top} = \Ib_{N},
\end{eqnarray}
where $\Ib_N$ denotes the $N\times N$ identity matrix.
Then, we can obtain the following matrix equality:
\begin{eqnarray}\label{eq:equiv}
\hank_{d|P}\left(\widehat \Zb \right) &=& \widetilde\Phib \Phib^{\top}\hank_{d|P}\left(\widehat \Zb \right) \Psib \tilde \Psib^{\top} =  \widetilde\Phib \Cb \tilde \Psib^{\top} \\
&=& \sum_{k=1}^M\sum_{l=1}^Q [\Cb]_{kl}\widetilde \Bb^{kl}
\end{eqnarray}
with $ [\Cb]_{kl}$ denoting the $(k,l)$-element of $\Cb \in \Cd^{M\times Q}$,
where 
\begin{eqnarray}\label{eq:C}
\Cb = \Phib^{\top}\hank_{d|P}\left(\widehat\Zb\right) \Psib 
\end{eqnarray}
is the so-called convolution framelet coefficient, and
\begin{eqnarray}\label{eq:B}
\widetilde\Bb^{kl} = \widetilde\phib_k\widetilde\psib_l^\top,\quad k=1,\cdots, M, ~l=1,\cdots, Q
\end{eqnarray}
refers the matrix basis that decomposes the Hankel matrix.

\begin{figure}[!tb]
\centering
\includegraphics[width=1.05\linewidth]{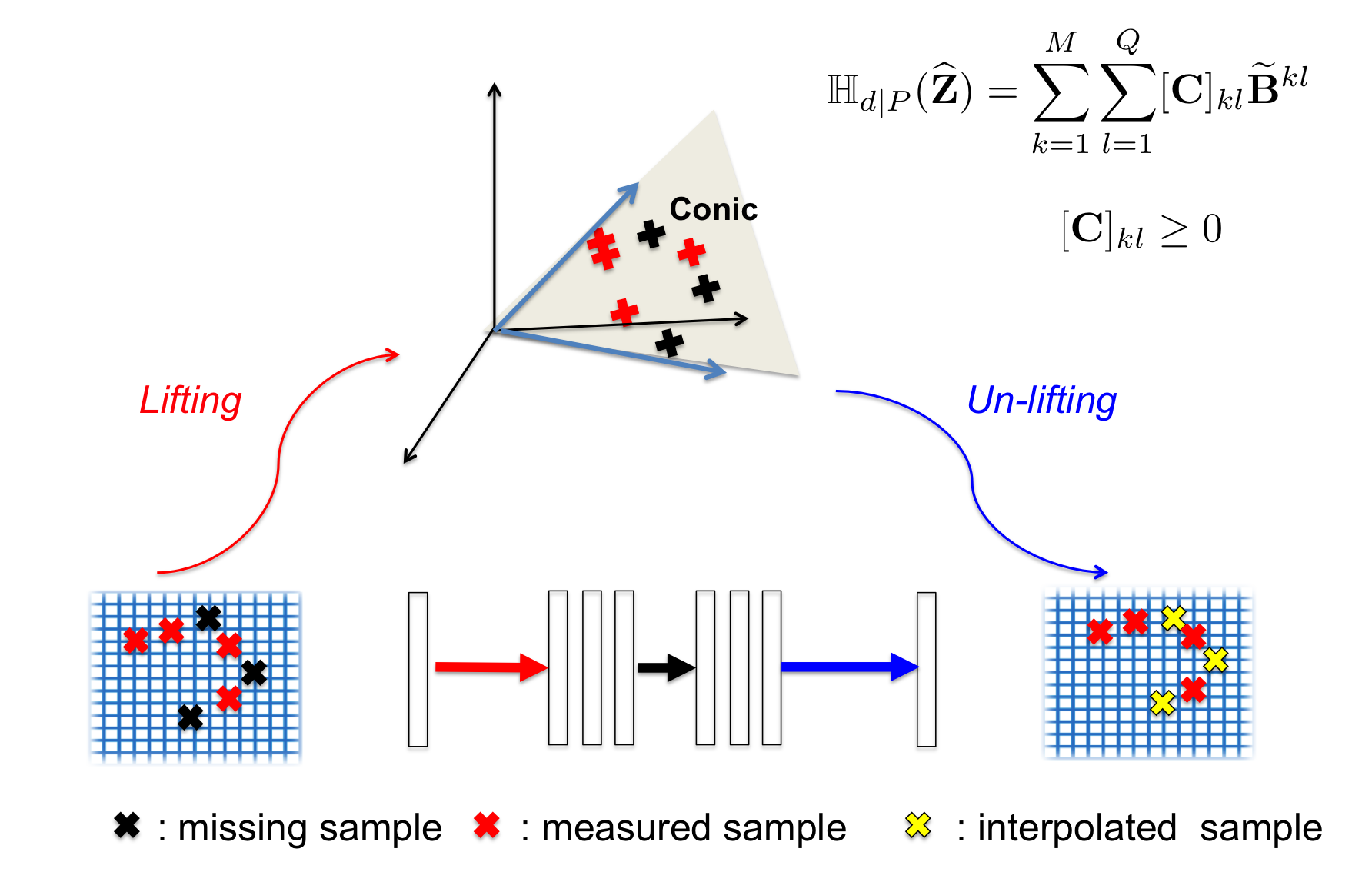}
\caption{Geometry of single layer encoder decoder network. The signal is first lifted into higher dimensional space, which is then decomposed into the positive combination of
bases. During this procedure, the missing $k$-space data (black color) are placed inside of the conic hull of the bases, so that they can be  interpolated
during the recomposition with conic bases. When this high dimensional conic decomposition procedure is observed in the original signal space, it becomes one level encoder-decoder neural network with ReLU.
}
\label{fig:geometry}
\end{figure}

One of the most important discoveries in \cite{ye2018deep} is that if
this  high-dimensional operation  is un-lifted to  the original signal space, it becomes
a single layer neural network with encoder-decoder architecture:
\begin{eqnarray} \label{eq:decomp0}
\Cb =  \Phib^\top \left(  \widehat \Zb \circledast \overline \Psib\right) , ~\quad
\widehat\Zb = % \hank_d^\dag\left( \hank_d(f) \right) %&=& \frac{1}{d} \sum_{i=1}^{m} \left( \tilde \Phi c_j\right) \circledast \tilde \psi_j \notag\\
 \left(\widetilde\Phib \Cb\right) \circledast \nu(\tilde \Psib),  
\end{eqnarray}
where $\circledast$ is the multi-channel input multi-channel output convolution, and
the convolutional filters of encoder and decoder layers are respectively given as follows:
\begin{align} \label{eq:enc_dec_filter}
&\overline\Psib:=
\begin{pmatrix}
\overline\psib_{1} & \cdots & \overline\psib_{Q}
\end{pmatrix}
 \in \Cd^{d \times r}, ~~\quad
\nu(\widetilde{\Psib}):=
\begin{pmatrix}
\widetilde{\psib}_{1} 
\\
\vdots 
\\
\widetilde{\psib}_{Q} 
\end{pmatrix} \in \Cd^{dr \times 1}.
\end{align}
where $\overline\psib_i$ denotes the flipped version of the vector $\psib_i$.

The main advantage of this explicit representation of the low-order Hankel matrix decomposition by the encoder-decoder structure is that it enables filter learning from the training data.
In particular, to prevent the learned bases from being significantly different from the training data, we enforce the following positivity constraint on the framelet coefficients:
$$[\Cb ]_{kl} \geq 0,~\forall k,l $$
Thus, the signal should live in the conical hull of the learned bases, so that learned bases and the signals are forced to live in geometric proximity (see Fig.~\ref{fig:geometry}).
Interestingly,  this conic (nonnegative) coding scheme is known
 to allow part-by representation, which is the key idea of   non-negative matrix factorization (NMF) \cite{
lee1997unsupervised,lee1999learning,lee1999learning,lee2001algorithms}.
This positivity constraint can be implemented using rectified linear unit (ReLU) during training.

Under this constraint,  $(MC')$ can be converted to the following equivalent problem:
\begin{eqnarray} 
\label{eq:image_regression2}
\quad \quad\quad \min_{\widehat{\Zb} \in \Hbc^0 }  && \sum_{i=1}^P\left\|g_i- \Fc^{-1}[\widehat{\zb}_i] \right\|^2 \\  
\mbox{subject to}%&& \rank~ \hank_{d|P} (\widehat \Zb) = Q \label{eq:sol} \\
&&\Pc_\Lambda[\widehat\gb_i ] = \Pc_\Lambda[\widehat \zb_i]  ,\quad i=1,\cdots, P \nonumber  \  .
\end{eqnarray}
%So far, the convolutional framelet expansion is  a linear representation, so we define a constraine signal space
 where $\Hbc^0$ denotes a constrained signal space:
\begin{eqnarray}
\Hbc^0 &=& \left\{  \Gb  \in \Rd^{N} \,\Big|\,\  \Gb = \left(\tilde\Phib   \Cb \right) \circledast \nu(\tilde \Psib  ), \right.  \notag \\
 && \left. \Cb  = \Phib^{\top} \left( \Gb \circledast \overline \Psib   \right) ,\right. \notag\\
 && \left.   [\Cb ]_{kl} \geq 0,~\forall k,l \right\} \notag  ,%\label{eq:finsuf}
 \end{eqnarray}
 where the convolution framelet coefficients $ [\Cb ]_{kl}$ are enforced to be non-negative.
Then, 
%the goal of the neural network training is to estimate the filters. % we are  given training data set
%Specifically, 
for a given $P$-channel training data set  $\{\widehat\yb_i^{(t)}, g_i^{(t)}\}_{i,t=1}^{P,T}$, where  $\widehat \yb_i^{(t)}$ and $g_i^{(t)}$ denotes the $t$-th batch 
under-sampled $k$-space data and the corresponding ground-truth image, respectively, from  the $i$-th coil,
%Then, from \eqref{eq:image_regression2}, 
the network training problem  can be formulated as follows:
\begin{eqnarray}\label{eq:newcost}
  \min_{ \Psib, \widetilde\Psib\in \Rd^{2d \times Q}}  \sum_{i=1}^P\sum_{t=1}^T\left\|g_i^{(t)}- \Fc^{-1}\Kc(\widehat\yb_i^{(t)};\Psib,\widetilde\Psib)\right\|^2 ,
\end{eqnarray}
Here,  the operator $\Kc: \Cd^{N} \to \Cd^{N}$ denotes the encoder-decoder network.
%and the filter $ \Psib, \widetilde\Psib$ are composite filters given by \eqref{eq:psi} and \eqref{eq:tpsi}, respectively.

The geometric implication of this training procedure for $k$-space interpolation is illustrated in
Fig.~\ref{fig:geometry}, where 1-D single coil  $k$-space data is used as an example for simplicity.
Specifically,  the original $k$-space data with missing element is  first {\em lifted} to higher dimensional space via Hankel matrix, 
which is then decomposed using the matrix bases $\widetilde\Bb_i^{kl}$ in \eqref{eq:B}.
Here, the training goal is to find the conic bases such that the measured and missing $k$-space data can
be represented as the conic (nonnegative) combination of the resulting basis so that the interpolation can be readily done during
the signal recomposition step using the bases.
When this conic coding procedure is unlifted to the original lower dimensional space,  the interpolated $k$-space data 
appear.  When this lifting, conic decomposition and unlifting
procedure are viewed from the original $k$-space domain, 
it becomes  one level encoder-decoder neural
network with ReLU.  In other word, an encoder-decoder network can be understood as a signal space manifestation of the 
conic coding of the signal being lifted to  a higher-dimensional space.

\begin{figure}[!t] 	
\center{ 
\includegraphics[width=9.cm]{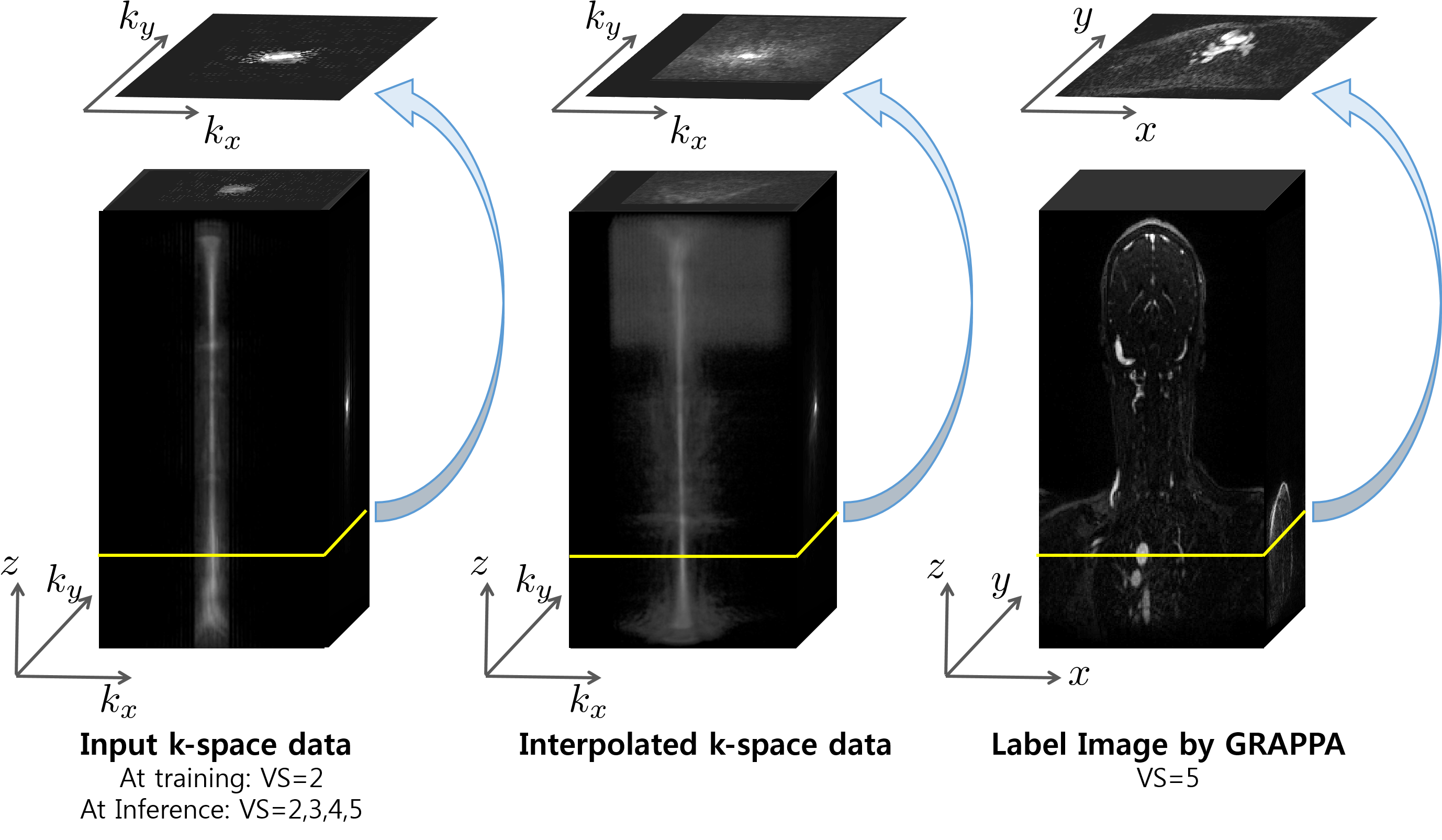}
}
\caption{Coordinate system for the data set. Here, VS stands for the number of view-sharing.}
\label{fig:coordinate}
\end{figure}

The idea can be further extended to the multi-layer deep convolutional framelet expansion, when the encoder and decoder convolution filter $\overline\Psib, \nu(\tilde\Psib) \in \Rd^{d\times Q}$
can be represented in a cascaded convolution of small length filters. For more details, see \cite{han2018k}.

\begin{figure*}[!hbt] 	
\center{ 
\includegraphics[width=16cm]{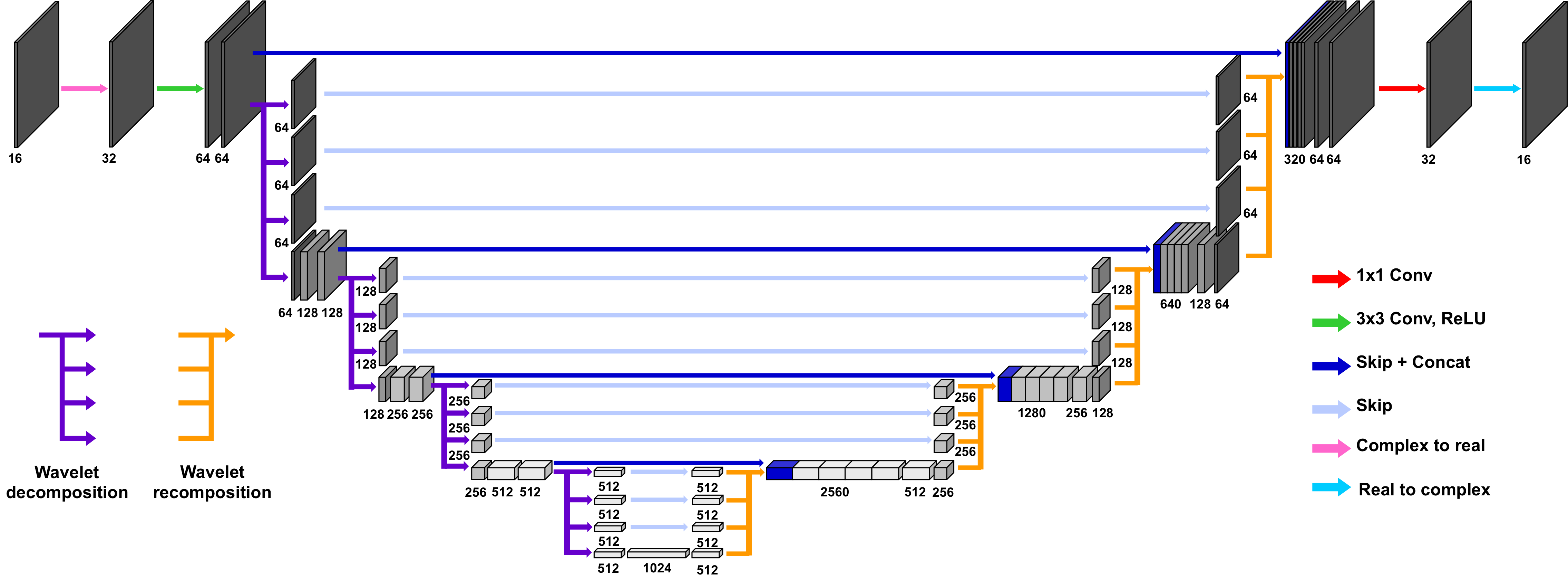}
}
\caption{ Network architecture of tight-frame U-net.}
\label{fig:network}
\end{figure*}

\subsection{Sparsification}

Many MR images can be sparsified using the finite difference
 \cite{jin2016general}. 
 In this case,  we can easily see that the extended Hankel matrix from the weighted $k$-space data given by
% $\hat z (\kb) = \hat h(\kb)\hat x (\kb)$ are low-ranked, 
\begin{eqnarray}
\begin{bmatrix} \hank_d(\widehat \hb\odot\hat \gb_1) &\cdots &  \hank_d(\widehat \hb\odot\hat \gb_P)\end{bmatrix}
\end{eqnarray}
has lower rank than \eqref{eq:PHankel}, where
 $\odot$ refers to the element-wise multiplication and the weight $\widehat\hb$ is  given by \cite{jin2016general,ye2017compressive}:
  \begin{eqnarray}\label{eq:h}
  \widehat\hb  = \begin{bmatrix}\widehat h(\kb_1) & \cdots & \widehat h(\kb_N)\end{bmatrix}^T \in \Cd^{N} 
  \end{eqnarray}
  with 
  \begin{eqnarray} %\quad\mbox{with}\quad
 \widehat h(\kb) := \Fc[h](\kb) =  \sin(\pi|\kb|),\quad  |\kb|\leq \frac{1}{2} \ .
 \end{eqnarray}
  Thus, the deep neural network is applied to the weighted $k$-space data to estimate
  the missing  spectral data $\widehat  h(\xb)\widehat g_i(\kb)$, after which the original $k$-space data is obtained
  by dividing with the same weight.
This can be easily implemented using a weighting and unweighting layer as shown in Fig.~\ref{fig:scheme}.

\section{Method}

\subsection{Training dataset}
Four sets of in vivo 3D DCE data were obtained with TWIST sequence using Siemens 3T Verio scanners in Gachon University Gil Medical Center. The data sets were for carotid vessel scan. The scanning parameters for two sets of carotid vessel data were as follows: repetition time (TR) = 2.5 ms, echo time (TE) = 0.94 ms, 159$\times$640$\times$80 matrix, 1.2 mm slice thickness, 16 coils, 30 temporal frames. For other two sets of carotid vessel data, the acquisition parameters were same as above, except for 2.5 mm slice thickness and 37 temporal frames. The sampling pattern of these data sets is described in Fig. \ref{fig:TWIST_concept}(a), where 24$\times$24 ACS regions were required for the conventional 2D GRAPPA reconstruction. In addition, the partial Fouirer was applied to the data, so only 63$\%$ of data was acquired. The downsampling rate was 3 and 2 along $k_x$ and $k_y$ direction, respectively. The read-out direction is $k_z$, which is fully sampled  (see Fig.~\ref{fig:coordinate}).
Then,  the input  $k$-space data for the neural network in Fig.~\ref{fig:scheme} is the $k_x - k_y$  data from the 
$k_x-k_y-z$ volume  in Fig.~\ref{fig:coordinate}, which is applied for each slice  along the readout direction and the temporal frames. 

Among four  patient  data sets of TWIST acquisition, we used three patient data for training and validation. The remaining one patient data was used for testing. More specifically, we used 33,810 slices for training and validation, and 12,210 slices for test.
%Since TWIST imaging aims at providing temporal information by highly accelerated acquisition, there are not fully acquired $k$-space data which can be used as label data to train the network. Therefore, we used the reconstructed images using GRAPPA and ALOHA as label data.

\subsection{Network architecture}

We now describe the CNN block in Fig.~\ref{fig:scheme}.
Note that the multi-channel convolution in  Eq.~\eqref{eq:decomp0} is complex-valued convolution.
Thus, to  convert the complex-valued multi-channel convolution to a real-valued ones,   we divide the complex-valued $ k $ space data into real and imaginary channels similar to \cite{han2018k}.
So, the actual implementation of Eq.~\eqref{eq:decomp0}  is as follows.  
First, the $P$-channel multi-coil $k$-space data $\widehat\Zb $ are  splitted into $2P$-channel input after  each $k$-space data has been split into real and image components.
Then, the encoder filters generates $Q$-channel outputs from these channel inputs using multi-channel convolution, after which
the pooling operation defined by $\Phib^\top$ is applied to each $Q$-channel output.
The resulting $Q$-channel feature maps corresponds to the convolutional framelet coefficients
(if there are multiple layers, this procedure is applied recursively).
Now, at the decoder, the $Q$-channel feature maps are processed using unpooling layer represented by $\tilde\Phib$,
which are then
 convoluted with the decoder filters to generate $P$-set of real and imaginary channels of the estimated $k$-space data.
 Finally,  complex valued $k$-space data  are  formed from each real and image channels. By doing this, all the successive
 layers are implemented using real-valued machine learning toolboxes.

In our prior work \cite{ye2018deep}, we also showed that the deep network with the encoder-decoder architecture mainly differ in their implementation
of the pooling and unpooling layers.
Here, we employed the tight-frame U-net \cite{ye2018deep}, which has Haar wavelet decomposition and recomposition as pooling and unpooling layers.
This choice of pooling and unpooling satisfies the so-called frame condition  \cite{ye2018deep},
 preserving  the detail of images. 
 This condition still holds even when we add additional by-pass connection \cite{ye2018deep}.

 For example, as shown in Fig. \ref{fig:network}, tight frame U-net was composed of convolution, batch nomarlization \cite{ioffe2015batch}, ReLU \cite{nair2010rectified}, and skip connection with concatenation. Each stage is composed of three $3 \times 3$ convolution layers followed by batch normalization and ReLU, except for the layer presented as a red arrow in Fig. \ref{fig:network}, which is $1 \times 1 $ convolution layer. 
In contrast to the standard U-net \cite{ronneberger2015u}, $2 \times 2$ pooling and unpooling were replaced by 2-D Haar wavelet decomposition and recomposition. Specifically, wavelet decomposition results in four subband (LL, LH, HL, HH bands), from which the LL band is only processed by following convolution layers, whereas the remaining subbands are skipped for wavelet recomposition. The number of convolutional filters increases from 64 in the first stage to 1024 in the final stage. 
Note that we have extra layers at the input and the output to transform complex-valued $k$ space data into real and imaginary channels, and vice versa.
Thus, the total number of channels for our tight-frame U-net is 32.

Since all data sets are from multi-channel acquistion, the square root of sum of squares (SSoS) images were finally applied after the coil-by-coil reconstruction. In addition, subtracted maximum intensity projection (MIP) to the SSoS images  was obtained to focus on the dynamics of the contrast agent.

\begin{figure*}[!bt] 	
\centerline{
\includegraphics[width=18cm]{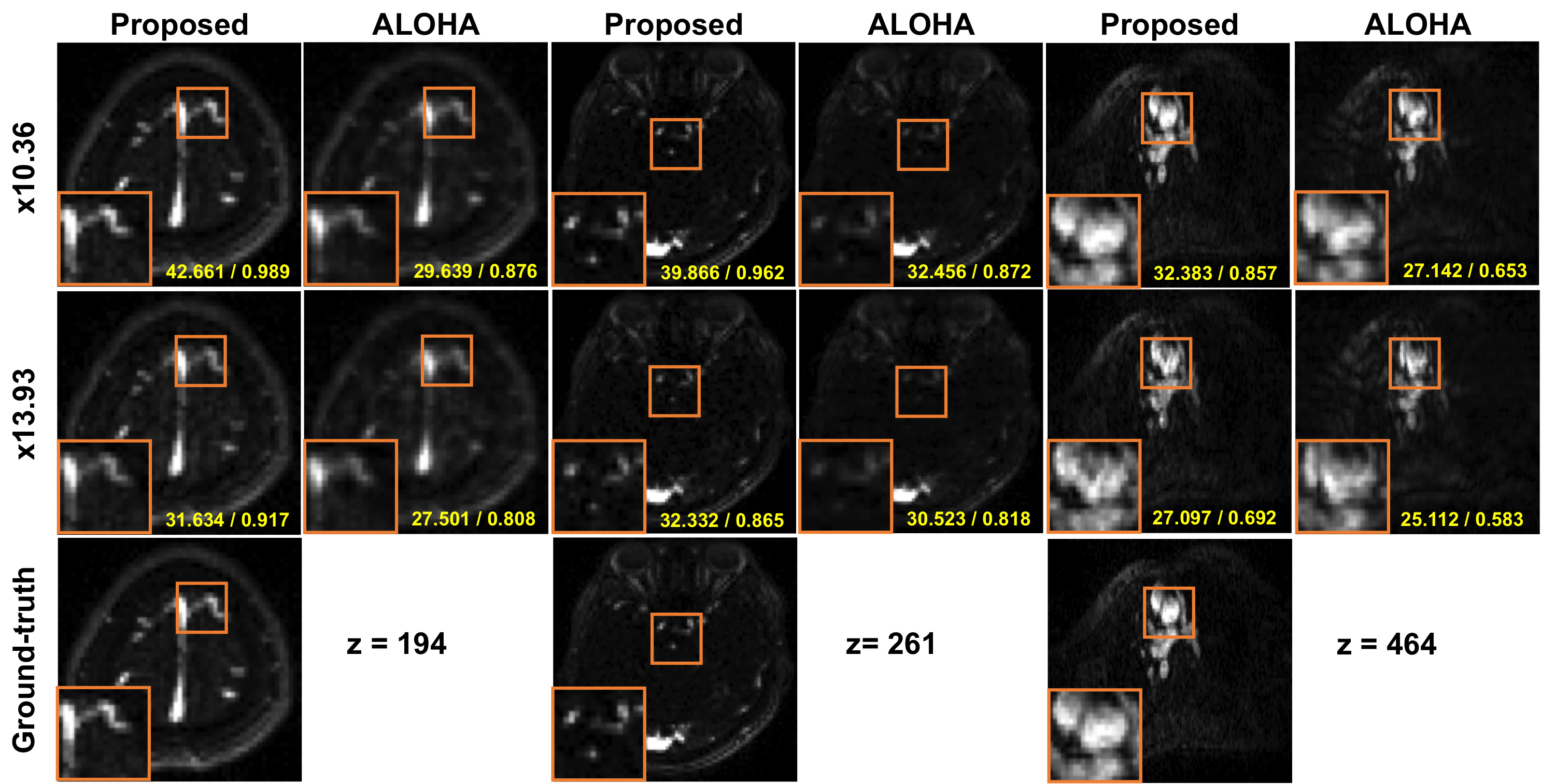}
}
\centerline{\mbox{(a)}}
\centerline{
\includegraphics[width=18cm]{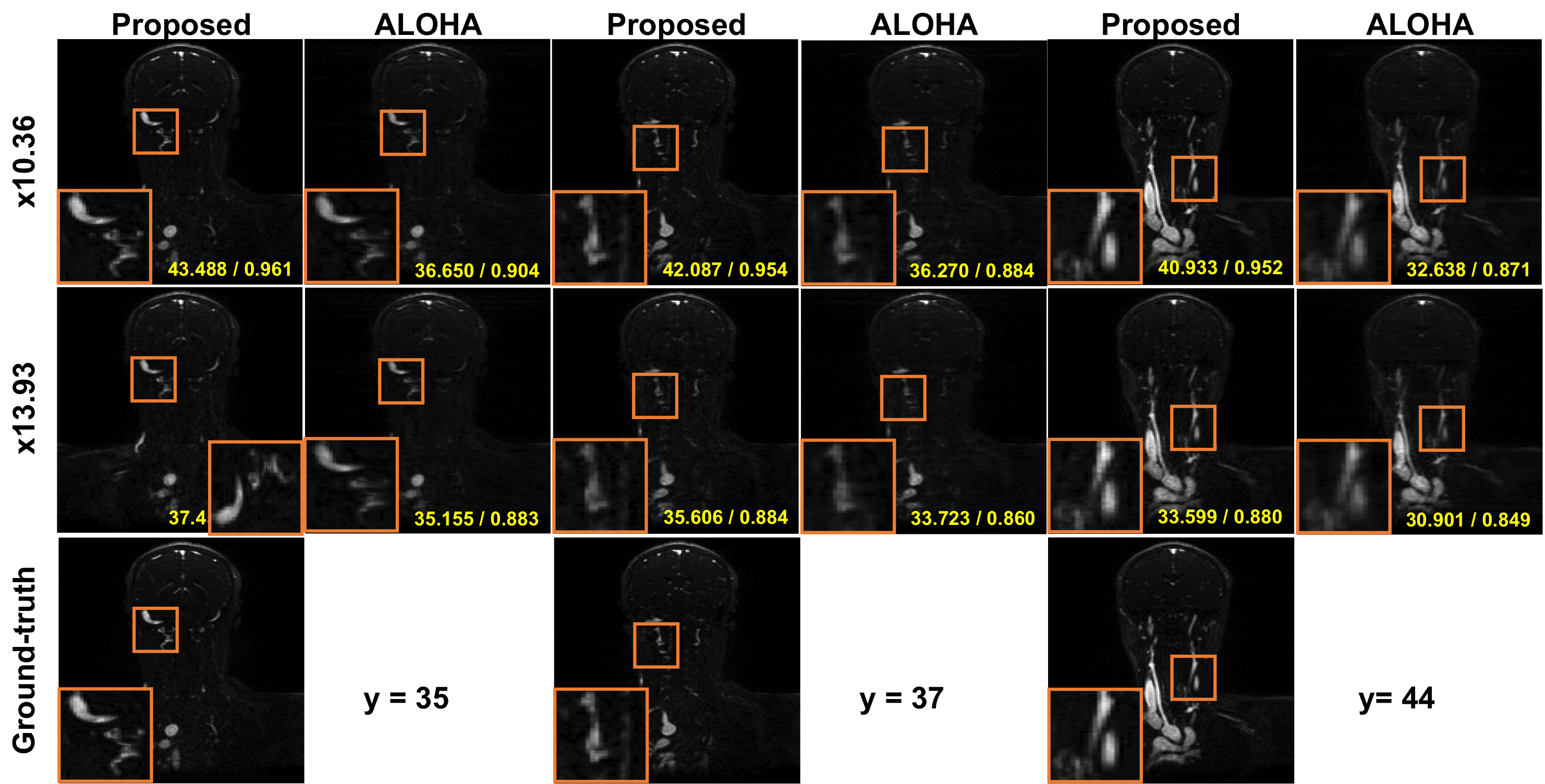}
}
\centerline{\mbox{(b)}}
\caption{ (a) Axial and  (b) coronal view of reconstruction results at various  downsampling factors. } 
\label{fig:all}
\end{figure*}

\begin{figure*}[!hbt] 	
\center{ 
\includegraphics[width=18cm]{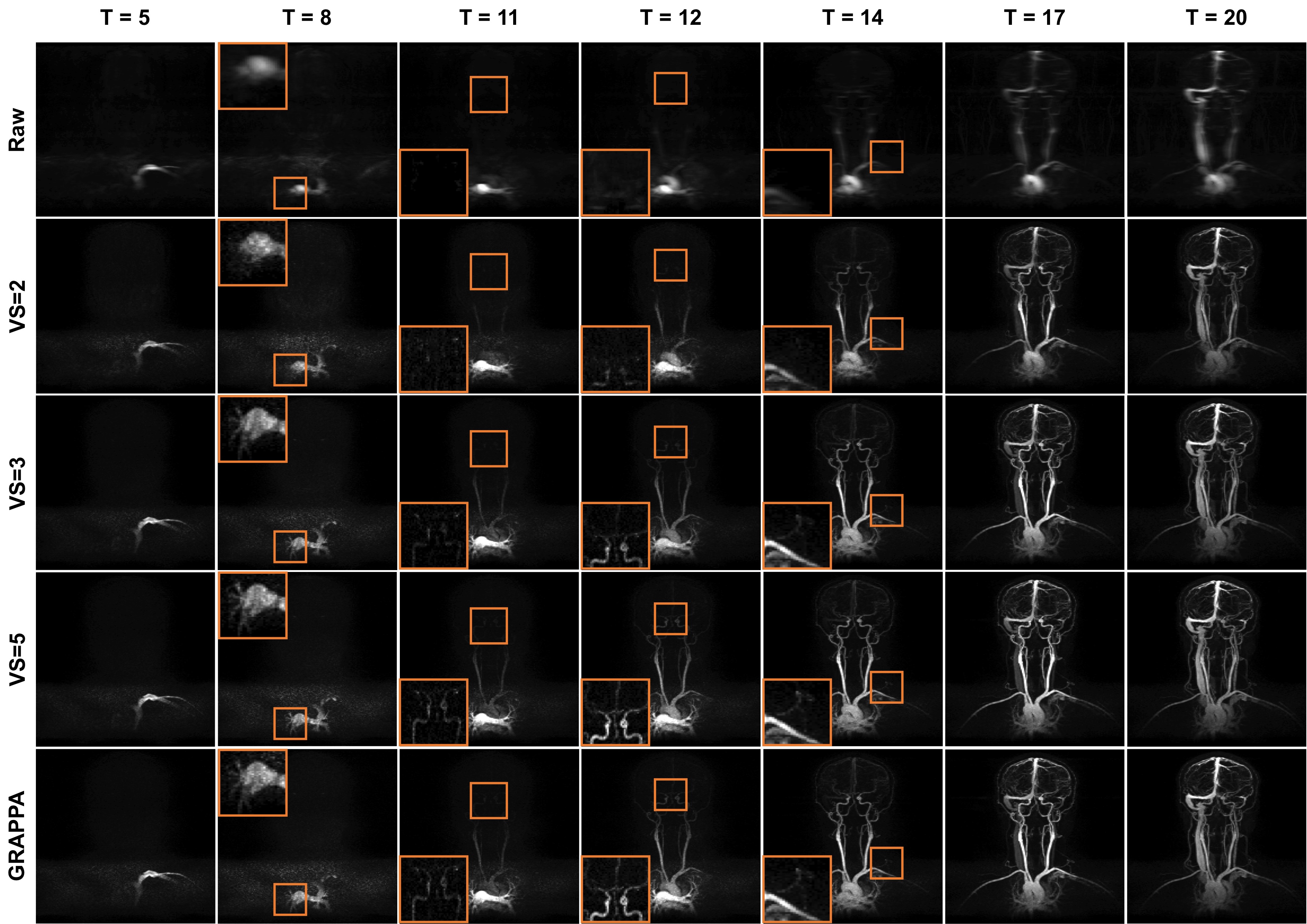}
}
\caption{Time resolution comparison of the reconstruction results of GRAPPA, raw data and the proposed methods for different view-sharing numbers. Here VS stands for the number of view sharing.}
\label{fig:temporal_resol}
\end{figure*}

\subsection{Network training}

%Since we do not have any ground-truth data, 
%For high sptio-temporal TWIST images, 

For our network training, we use the $k$-space data from the two adjacent view sharing as input (see Fig.~\ref{fig:TWIST_concept}(b) and Fig.~\ref{fig:coordinate}),
whereas  the coil-by-coil  reconstructed images using GRAPPA with five view-sharing are used  as labels.
%Specifically, our neural network is trained in two different ways. First,  the
%neural network is trained using only GRAPPA data as labels.  
%This network is used to evaluate the spatial resolution
%of the proposed  $k$-space deep learning  compared to other CS algorithm such as ALOHA.
The input data for the inference stage are the downsampled $k$-space data using 2 to 5 contiguous frames, so that we can produce the images with various temporal resolution.
%
% the spatial resolution of this network is the maximum,  the temporal resolution is determined by the GRAPPA reconstruction.
%Thus, our second neural network is trained using both GRAPPA and ALOHA reconstruction.
%More specifically, we first trained the network using GRAPPA reconstruction as labels. 
% After the learning converged, the label data for training was replaced with the reconstructed $k$-space using ALOHA.
%Using this fine-tuning, our goal is to improve the temporal resolution.

 The parameters for GRAPPA  reconstruction were chosen to provide the best results. The kernel size for GRAPPA is 5$\times$5 for the data sets. 
As a representative CS reconstruction method, the ALOHA \cite{jin2016general} was used.
 The parameters for ALOHA were as following: annihilating filter size = 13$\times$5 , 3 levels of pyramidal decomposition, decreasing LMaFit tolerance values ($10^{-3}, 10^{-4}, 10^{-5}$) at each level, and ADMM parameter $\mu$ = $10^{-1}$.

The  network   was trained using Adam optimization \cite{kingma2014adam} with the momentum $\beta _1 = 0.9$ and $\beta_2 = 0.999$. We used $l_2$ loss in the image domain. The initial learning rate was $10^{-2}$, and it was divided in half at every 50 epochs. The size of mini-batch was 40. The number of epochs to train this pre-trained network was 150. 
%Using given pre-trained network, fine-tuning was performed using the reconstructed images using ALOHA for improvement of temporal resolution. 
%The parameters for fine-tuned the network  were same as those for pre-trained network except for the learning rate. The initial learning rate was set to $10^{-4}$ and decreased in same way as the pre-trained network. We used 100 epochs for training using ALOHA reconstruction data. 

The proposed network was implemented in Python using TensorFlow library \cite{abadi2016tensorflow} and trained using an NVidia GeForce GTX 1080-Ti graphics processing unit.
It took about 6 days for the network training.

\subsection{Comparative Studies}

To evaluate the performance of the proposed method,  the peak signal-to-noise ratio (PSNR) and structural similarity (SSIM) index \cite{wang2004image} were calculated as  quantitative metrics. Since the reconstruction outputs from the network are multi-coil images, we calculate the sum-of-squares images to combine all the coil images  to calculate
these quantitative metrics.
 The PSNR is defined as
\begin{eqnarray}
	PSNR %&=& 10 \cdot \log_{10} \left(\dfrac{MAX_{Y}^2}{MSE(\widehat{X}), Y)})\right) \\
		 &=& 20 \cdot \log_{10} \left(\dfrac{MAX_{Y}}{\sqrt{MSE(\widehat{X}, Y)}}\right), 
\label{eq:psnr}		 
\end{eqnarray}
where $\widehat{X}$ and $Y$ denote the reconstructed sum-of-squares image and noise-free sum-of-squares image (ground truth), respectively. $MAX_{Y}$ is the maximum value of noise-free sum-of-squares image.
SSIM is used to measure the similarity between original image and distorted image due to deformation, and it is defined as
\begin{equation}
	SSIM = \dfrac{(2\mu_{\widehat{X}}\mu_{Y}+c_1)(2\sigma_{\widehat{X}Y}+c_2)}{(\mu_{\widehat{X}}^2+\mu_{Y}^2+c_1)(\sigma_{\widehat{X}}^2+\sigma_{Y}^2+c_2)},
\end{equation}
where $\mu_{M}$ is a average of $M$, $\sigma_{M}^2$ is a variance of $M$ and $\sigma_{MN}$ is a covariance of $M$ and $N$. 
To stabilize the division, $c_1=(k_1R)^2$ and $c_2=(k_2R)^2$ are defined in terms of $R$, which is the dynamic range of the pixel values. We followed the default values of $k_1 = 0.01$ and $k_2 = 0.03$.

\section{Results}\label{sec:result}

%\subsection{Parallel Imaging Performance}

We first compare the performance of the proposed  $k$-space deep learning for parallel MR image reconstruction. %with ALOHA-based compressed sensing approach. 
In this experiments,  the fully sampled $k$-space data  generated
using GRAPPA were considered as the ground-truth, from which retrospective sub-sampling  were performed at various downsampling ratios, which corresponds
to specific view sharing number (VS).
Then, the proposed multi-coil $k$-space deep learning and ALOHA method were applied for comparative studies. The axial reconstruction results  
and the coronal reformatted images in Fig.~\ref{fig:all} clearly show that the proposed method significantly outperformed  ALOHA.
For example, near perfect reconstructions are obtained at $R=10.36$ downsampling (i.e. VS=3) using the proposed method, while still blurry images
are observed from the ALOHA reconstruction.
In terms of PSNR and SSIM values,  the proposed method significantly outperforms the ALOHA.
For example, from the axial reconstruction images,
at $R=13.93$ acceleration factor (i.e. VS=2), the proposed $k$-space deep learning is about $2\sim 4$dB better than ALOHA in terms of PSNR,
whereas at $R=10.36$ downsampling (i.e. VS=3), the proposed $k$-space deep learning is about $5\sim 13$dB better than ALOHA.
When we compare the results in coronal reformatted images, similar PSNR gains were observed.
At $R=13.93$ acceleration factor, the proposed $k$-space deep learning is about $2\sim 3$dB better than ALOHA in terms of PSNR,
whereas at $R=10.36$, the proposed $k$-space deep learning is about $6\sim 8$dB better than ALOHA.

{
Reconstruction results of the carotid vessel data sets for test result are demonstrated in Fig. \ref{fig:temporal_resol}. % and Fig. \ref{fig:test_result}, respectively. 
The temporal frames were chosen to illustrate the propagation of the contrast agent and to compare the temporal resolution.
In the proposed method,  the same neural network  can produce reconstruction results using various
view sharing number,  so  we provide reconstruction results with VS=2, 3, and 5.
The raw data in Fig. \ref{fig:temporal_resol} were obtained by directly applying inverse Fast Fourier Transform (FFT) to the $k$-space data without view-sharing, which appear blurry but still  have the true temporal resolution. The raw data reconstruction results are used as for evaluating the temporal resolution.
By inspection, we can see that  the contrast agent was abruptly spread out in the GRAPPA reconstruction. For example, there was a rapid propagation of contrast agent from the $T=11$ frame to  the $T=12$ frame as shown in Fig. \ref{fig:temporal_resol}. This is because the combination of several temporal frames resulted in the blurring of the temporal resolution. Therefore, the flow of the contrast agent was suddenly changed between just one frame. This degradation of temporal dynamics could be found frequently as the number of view-sharing increased.
 In the reconstructed images using the proposed method, the flow of the contrast agent and the detail of dynamics were captured correctly with
 the VS=2, and a minor temporal blurring starts with VS=3.   With VS=5, which is the equal to the view sharing used in GRAPPA, the spatial and temporal resolution of the proposed methods were near identical to those of GRAPPA.
More specifically, the contrast agent shown in the GRAPPA reconstruction was not visible in the raw image, and the proposed reconstruction of VS = 2 to VS = 5 at $ T = 12 $ frame clearly shows the temporal resolution degradation with the increase in the number of view sharing.
In the GRAPPA reconstruction, the detail of the spread of the contrast agent was influenced by the high-frequency region in the $k$-space so that the temporal dynamics of the future frame was erroneously incorporated in the current frame. 
%Moreover, heart motion can be  better captured in the proposed reconstruction with VS=2 than in the GRAPPA reconstruction because the $k$-space data was interpolated using only the current and the previous frames in the proposed method, that is, the blurring of temporal resolution was significantly reduced in the proposed method.
 }

In addition, the proposed method is computationally more efficient than GRAPPA and ALOHA as shown in
Table~\ref{tab:time}.  Specifically, the proposed method is several order of magnitude faster than GRAPPA and ALOHA.

\begin{table}[!hbt]
\centering
\resizebox{0.45\textwidth}{!}{
	\begin{tabular}{c|ccc}
		\hline
				& GRAPPA & ALOHA   & Proposed \\ \hline\hline
		Time/slice  (sec) & 6.09 & 84.61& 0.029 \\ \hline
	\end{tabular}}
	\caption{Computational time for 16 coil $k$-space interpolation time using various methods.}
	\label{tab:time}
\end{table}

\section{Discussion}

Although we have mainly focused on the TWIST reconstruction, the proposed $k$-space deep learning approach
can be used for various parallel imaging applications.  Similar to the results in Fig.~\ref{fig:all}, we expect the
significant gain compared to the existing parallel imaging and  compressed sensing approaches.

%The GRAPPA kernel, calculated from ACS regions,  was used to reconstruct the $k$-space data. 

%We found that  GRAPPA reconstruction shows noise boosting when the ACS regions are not sufficiently large. 
%Therefore, the GRAPPA reconstruction had problems with noise boosting in the background. 
%The $k$-space weighting schemes in 
ALOHA improved the image quality for TWIST imaging \cite{7547372}, but we found that
the spatial resolution of ALOHA reconstruction  was not sufficiently good at high acceleration factor from the reduced view sharing.
On the other hand, the proposed method significantly improves the performance at high acceleration factor.
This may lead to an interesting question: why the proposed $k$-space deep learning outperforms ALOHA even though they are related to each other?

\begin{figure}[!t] 	
\center{
\includegraphics[width=7cm]{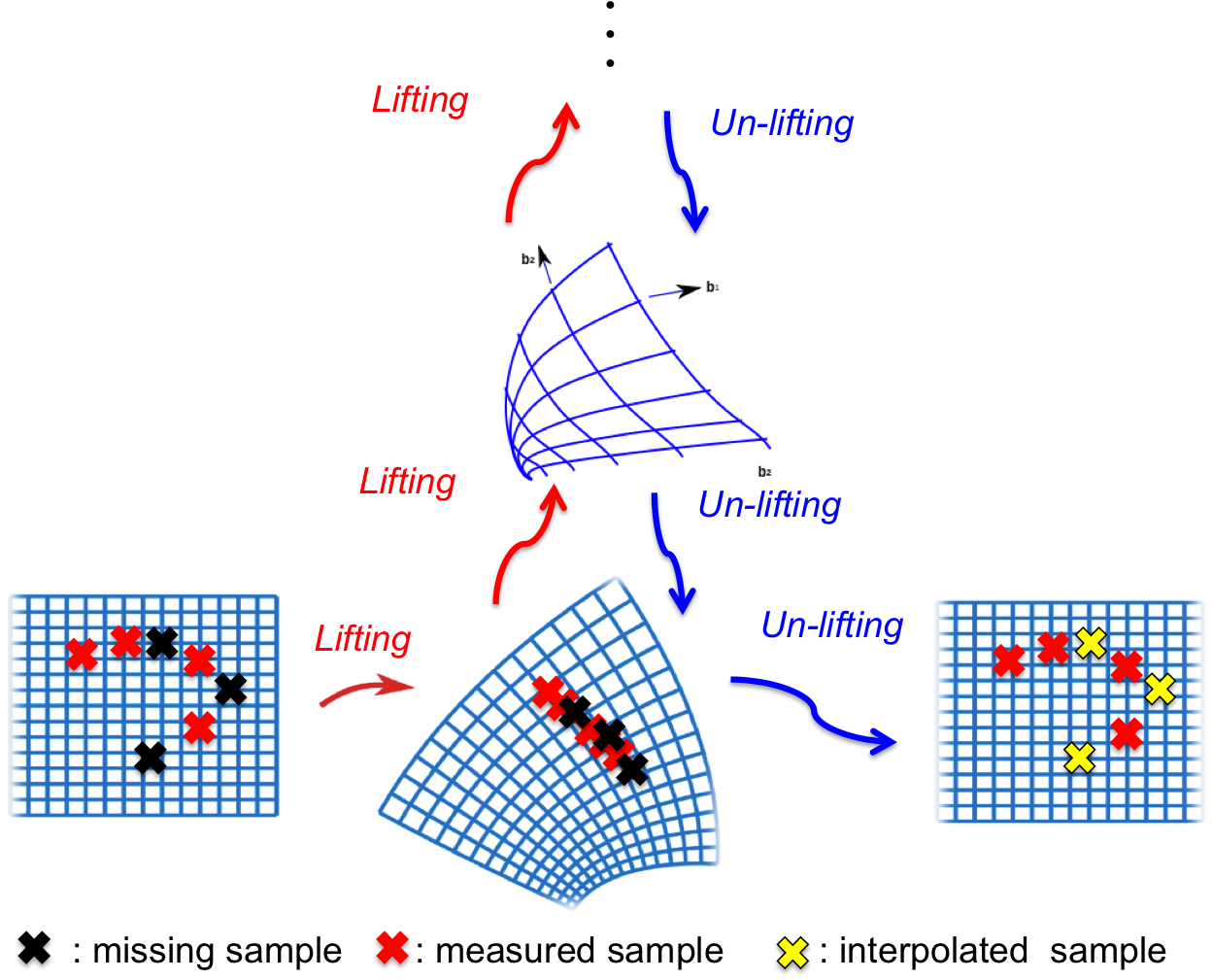}
}
\caption{Geometry of multi-layer encoder decoder architecture.} 
\label{fig:multilayer}
\end{figure}

It is important to note that the proposed neural
network is not another  implementation  of ALOHA for computational saving, 
but that it is a new algorithm that significantly improves the interpolation performance of ALOHA by exploiting the exponential expressiveness of a deep neural network
by increasing the number of layers.  In fact, our geometric view in Fig.~\ref{fig:geometry} is for the single layer encoder-decoder architecture, and the deep
neural network is obtained by recursively applying this high-dimensional lifting, conic coding, and un-lifting as shown in Fig.~\ref{fig:multilayer}.  During this recursive lifting to higher dimensional
space, it is believed that
the complicated input  $k$-space data  can have simpler manifold that can be exploited for better  interpolation.

As shown in the Table~\ref{tab:time}, the computational time of the proposed method is more than 100 times faster than the GRAPPA reconstruction because we do not need to compute the interpolation kernel from the data on the fly.
This is because the neural network has already learned the interpolation kernel from the training data, so all necessary calculations are simple convolution and pooling. The significant computational savings, in addition to the flexibility of obtaining reconstruction results at various numbers of view sharing, may suggest a new paradigm for DCE-MRI.
In particular, instead of using a fixed spatial and temporal resolution for dynamic studies, our methods can immediately generate reconstruction results for all possible combinations of spatial and temporal resolutions by simply changing the view-sharing number.
Thus, the proposed method may allow more accurate and quantitative time-resolved angiography studies.

Given the flexibility to generate images at various temporal resolution, one may wonder how thus-trained
neural network can be generalized to the unseen images with faster temporal resolution.
%One of the main reasons for the excellent generalization capability is 
Recall that our neural network is trained to learn
the structure of the weighted $k$-space data  based on the observation
that the Hankel matrix is low-ranked. Thus, the low-rankness is not a specific feature
from  
GRAPPA reconstruction; rather it is a general feature of $k$-space data from MR images. Therefore,
as long as our neural network is trained to learn the structure of the Fourier data, it can be generalized well to images at any intermediate time frame. %temporal resolution.
This is another important advantage of the proposed $k$-space deep learning.

%
% Since the network learned the mapping to GRAPPA reconstruction from pre-training, the more detail of image can be captured in the proposed reconstruction as shown in yellow boxed in Fig. \ref{fig:test_result}(a). This result verifies that the proposed method provided better spatial and temporal resolution.  

\section{Conclusion}\label{sec:conclusion}

The purpose of this study was to improve the temporal resolution of TWIST imaging
and to propose an algorithm that generates reconstruction results at various sliding window size.
To address this problem, we developed a novel $k$-space deep learning algorithm for parallel MRI.
Specifically, based on the recent mathematical discovery that a deep neural network
can be developed as a multilayer extension of data-driven Hankel matrix decomposition,
our $k$-space deep neural network was designed to simultaneously exploit the
multi-coil diversity and spatial domain redundancy.

The improvement of temporal resolution in TWIST imaging was verified by the reconstruction of in vivo data sets. Moreover,
it was demonstrated that one trained network can immediately generate multiple reconstruction results with various spatial and temporal resolution trade-off
by simply changing the number of view sharing at the inference stage.
As this method can
be used with the existing TWIST acquisition protocol without any modification of pulse sequence,  we believe that the method provides an important new research direction
that can significantly extend the clinical applications.

%
%%
%\bibliographystyle{IEEEtran}
%%\bibliographystyle{MRM}
%\bibliography{submit_bib,ref}
%%


\clearpage

%\clearpage
\section*{\Large SUPPLEMENTARY MATERIAL}
\section*{Appendix A}

For simplicity, here we consider 1-D signals, but its extension to 2-D is straightforward.
 In addition,  to avoid separate treatment of boundary conditions, we assume periodic boundary.
Let $\fb\in \Cd^N$ be the signal vector. 
Then, a single-input single-output (SISO) convolution of the input $\fb$ and the length $d$ filter $\overline \hb \in \Rd^d$  can be represented in a matrix form:
\begin{eqnarray}\label{eq:SISO}
\yb = \fb\circledast \overline\hb &=& \hank_d(\fb) \hb \ ,
\end{eqnarray}
where  $\hank_d(\fb)$ is a wrap-around  Hankel matrix  defined by
 \begin{eqnarray} %\label{eq:hank}
\hank_d(\fb) = %\left[
        \begin{bmatrix}
        [\fb]_1  &   [\fb]_2 & \cdots   &   [\fb]_d    \\
        [\fb]_2  &   [\fb]_3 & \cdots   &   [\fb]_{d +1}   \\
         \vdots    & \vdots     &  \ddots    & \vdots    \\
%        f[n-d+1]  &   f[n-d+2] & \cdots &   f[n]\\ \hline 
%        f[n-d+2]  &   f[n-d+3] & \cdots &   f[1] \\
%           \vdots    & \vdots     &  \ddots    & \vdots    \\
        [\fb]_N  &   [\fb]_1 & \cdots   &   [\fb]_{d -1}   %  &\cdots&   \yb_i(1)\\
        \end{bmatrix}
%    \right] , %\quad \quad\quad \in \Cd^{n\times d}  \  .
    \end{eqnarray}
where $d$ denotes the matrix pencil parameter.
On the other hand, multi-input multi-output (MIMO) convolution for the $P$-channel
input $\Zb=[\zb_1,\cdots,\zb_P]$ to generate $Q$-channel output
$\Yb=[\yb_1,\cdots,\yb_Q]$
can be represented  by
\begin{eqnarray}\label{eq:MIMO}
\yb_i = \sum_{j=1}^{P} \zb_j\circledast \overline\psib_i^j,\quad i=1,\cdots, Q
\end{eqnarray}
where 
$\overline\psib_i^j \in \Rd^d$ denotes the length $d$- filter that convolves the $j$-th channel input to compute its contribution to 
the
$i$-th output channel. 
By defining the MIMO filter kernel $\Psib$ as follows:
\begin{eqnarray}
\Psib = \begin{bmatrix} \Psib_1 \\ \vdots \\ \Psib_P \end{bmatrix} \, \quad \mbox{where} \quad \Psib_j =  \begin{bmatrix} \psib_1^j  & \cdots & \psib_Q^j \end{bmatrix} \in \Rd^{d\times Q}  
\end{eqnarray}
the corresponding matrix representation of the MIMO convolution is then given by 
\begin{eqnarray}
\Yb &=& \Zb \circledast \overline\Psib \label{eq:MIMO_form}\\
&=& \sum_{j=1}^P \hank_d(\zb_j) \Psib_j  \label{eq:multifilter0}\\
%&=& \hank_{d|p}\left(Z\right) \begin{bmatrix} \Psi_1 \\ \vdots \\ \Psi_p \end{bmatrix} \notag\\
&=& \hank_{d|P}\left(\Zb\right) \Psib \label{eq:multifilter}
\end{eqnarray}
where    $\overline\Psib$ is a flipped block structured matrix:
\begin{eqnarray}
\overline\Psib = \begin{bmatrix} \overline\Psib_1 \\ \vdots \\ \overline\Psib_P \end{bmatrix} \, \quad \mbox{where} \quad \Psib_j =  \begin{bmatrix} \overline\psib_1^j  & \cdots & \overline\psib_Q^j \end{bmatrix} \in \Rd^{d\times Q}  
\end{eqnarray}
and $\hank_{d|P}\left(\Zb\right)$ is  an {\em extended Hankel matrix}  by stacking  $P$ Hankel matrices side by side: %  given by
\begin{eqnarray}\label{eq:ehank}
\hank_{d|P}\left(\Zb\right)  := \begin{bmatrix} \hank_d(\zb_1) & \hank_d(\zb_2) & \cdots & \hank_d(\zb_P) \end{bmatrix} \ . %\quad \in f(n,d;p)
\end{eqnarray}

\end{document}